\documentclass[conference]{IEEEtran}

\IEEEoverridecommandlockouts

\usepackage{cite}
\usepackage{amsmath,amssymb,amsfonts}
\usepackage{algorithmic}
\usepackage{graphicx}
\usepackage{textcomp}
\usepackage[xindy,acronym]{glossaries}
\usepackage{xcolor}
\usepackage{comment}

\def\BibTeX{{\rm B\kern-.05em{\sc i\kern-.025em b}\kern-.08em
    T\kern-.1667em\lower.7ex\hbox{E}\kern-.125emX}}

\begin{document}

\newacronym{TV}{TV}{television}
\newacronym{DTTB}{DTTB}{digital television terrestrial broadcasting}
\newacronym{DVB}{DVB}{Digital Video Broadcast}
\newacronym{DVB-H}{DVB-H}{Digital Video Broadcast-Handheld}
\newacronym{ATSC}{ATSC}{Advanced Television System Committee}
\newacronym{ATSC-M/H}{ATSC-M/H}{Advanced Television System Committee - Mobile/Handheld}
\newacronym{IPTV}{IPTV}{Internet Protocol television}
\newacronym{IP}{IP}{Internet Protocol}
\newacronym{l1}{L1}{Layer 1}
\newacronym{l2}{L2}{Layer 2}
\newacronym{l3}{L3}{Layer 3}
\newacronym{cnn}{CNN}{Convolutional Neural Network}
\newacronym{ann}{ANN}{Artificial Neural Network}
\newacronym{scef}{SCEF}{Service Capability Exposure Function}
\newacronym{zipnet}{ZipNet}{Zipper Network}
\newacronym{rrm}{RRM}{Radio Resource Management}
\newacronym{fps}{FPS}{Frames Per Second}
\newacronym{qa}{QA}{Question Answering}
\newacronym{ar}{AR}{Augmeneted Reality}

\newacronym{l1}{L1}{Layer 1}

\newacronym{cdl}{CDL}{Clustered Delay Line}
\newacronym{tdl}{TDL}{Tapped Delay Line}

\newacronym{HR}{HR}{Human Resources}

\newacronym{CSI}{CSI}{Channel State Information}
\newacronym{ml}{ML}{Machine Learning}
\newacronym{bss}{BSS}{Business Support System}
\newacronym{nlp}{NLP}{Natural Language Processing}
\newacronym{5g}{5G}{Fifth Generation of Mobile Networks}
\newacronym{6g}{6G}{Sixth Generation of Mobile Networks}
\newacronym
[
  longplural={Large Language Models}
]
{llm}{LLM}{Large Language Model}
\newacronym
[
  longplural={Bidirectional Encoder Representations from Transformers}
]
{bert}{BERT}{Bidirectional Encoder Representations from Transformer}
\newacronym
[
  longplural={Universal Sentence Encoder}
]
{use}{USE}{Universal Sentence Encoder}
\newacronym
[
  longplural={Radio Base Stations}
]
{rbs}{RBS}{Radio Base Station}
\newacronym
[
  longplural={Natural Language Understanding}
]
{nlu}{NLU}{Natural Language Understanding}
\newacronym
[
  longplural={Radio Base Stations}
]
{nf}{NF}{Network Function}
\newacronym
[
  longplural={Generative Pretrained Transformers}
]
{gpt}{GPT}{Generative Pretrained Transformer}
\newacronym
[
  longplural={Generative Pretrained Transformers}
]
{rouge}{ROUGE}{Recall-Oriented Understudy for Gisting Evaluation}
\newacronym
[
  longplural={Deep Belief Networks}
]
{dbn}{DBN}{Deep Belief Network}
\newacronym
[
  longplural={Boltzmann Machines}
]
{bm}{BM}{Boltzmann Machine}
\newacronym
[
  longplural={Variational Autoencoders}
]
{vae}{VAE}{Variational Autoencoder}
\newacronym
[
  longplural={Long-Short Term Memory networks}
]
{lstm}{LSTM}{Long-Short Term Memory network}
\newacronym
[
  longplural={Probability Density Functions}
]
{pdf}{PDF}{Probability Density Function}
\newacronym
[
  longplural={Recurrent Neural Networks}
]
{rnn}{RNN}{Recurrent Neural Network}
\newacronym
[
  longplural={Generative Adversarial Networks}
]
{gan}{GAN}{Generative Adversarial Network}
\newacronym
[
  longplural={Restricted Boltzmann Machines}
]
{rbm}{RBM}{Restricted Boltzmann Machine}
\newacronym
[
  longplural={Digital Twins}
]
{dt}{DT}{Digital Twin}

\newacronym{itu}{ITU}{International Telecommunication Union}
\newacronym{etsi}{ETSI}{European Telecommunication Standards Institute}
\newacronym{api}{API}{Application Program Interface}
\newacronym{ue}{UE}{User Equipment}
\newacronym{PC}{PC}{personal computer}
\newacronym{RAN}{RAN}{radio access network}
\newacronym{CN}{CN}{core network}
\newacronym{MS}{MS}{mobile station}
\newacronym{soa}{SoA}{state of the art}

\newacronym{genai}{GAI}{Generative AI}

\newacronym{ITU-R}{ITU-R}{International Telecommunications Union - Radiocommunication Sector}
\newacronym{IMT-Advanced}{IMT-Advanced}{International Mobile Telecommunications Advanced}
\newacronym{4G}{4G}{fourth-generation of mobile phone communications and Internet access technology}

\newacronym{rlhf}{RLHF}{Reinforcement Learning with Human Feedback}
\newacronym{3gpp}{3GPP}{Third Generation Partnership Project}
\newacronym{GSM}{GSM}{Global System for Mobile Communications}
\newacronym{UMTS}{UMTS}{Universal Mobile Telecommunications System}
\newacronym{HSPA}{HSPA}{High Speed Packet Access}
\newacronym{lte}{LTE}{Long-Term Evolution}
\newacronym{lte-a}{LTE-A}{Long-Term Evolution Advanced}

\newacronym{e-UTRAN}{e-UTRAN}{evolved Universal Terrestrial Radio Access Network}
\newacronym{eNB}{eNB}{e-UTRAN NodeB}
\newacronym{gNB}{gNB}{gNodeB}
\newacronym{EPC}{EPC}{Evolved Packet Core}

\newacronym{MBMS}{MBMS}{Multimedia and Broadcast Multicast Service}
\newacronym{eMBMS}{eMBMS}{Evolved MBMS}
\newacronym{SFN}{SFN}{single-frequency network}
\newacronym{MBSFN}{MBSFN}{MBMS single-frequency network}
\newacronym{BM-SC}{BM-SC}{Broadcast/Multicast Service Center}
\newacronym{MBMS GW}{MBMS GW}{MBMS Gateway}
\newacronym{MME}{MME}{Mobility Management Entity}
\newacronym{MCE}{MCE}{Multi-cell/multicast Coordinating Entity}
\newacronym{SYNC}{SYNC}{synchronization}
\newacronym{MCCH}{MCCH}{Multicast Control Channel}
\newacronym{MTCH}{MTCH}{Multicast Traffic Channel}
\newacronym{MCH}{MCH}{Multicast Channel}
\newacronym{PMCH}{PMCH}{Physical Multicast Channel}
\newacronym{PDSCH}{PDSCH}{Physical Downlink Shared Channel}
\newacronym{tmforum}{TMForum}{TeleManagement Forum}

\newacronym{sop}{SOP}{Sentence Order Prediction}
\newacronym{fid}{FID}{Fréchet Inception Distance}
\newacronym{is}{IS}{Inception Score}
\newacronym{mlm}{MLM}{Masked Language Modeling}
\newacronym{nsp}{NSP}{Next Sentence Prediction}
\newacronym{blue}{BLUE}{Bilingual Evaluation Understudy}
\newacronym{cer}{CER}{Concept Error Rate}

\newacronym{hss}{HSS}{Home Subscriber Server}

\newacronym{sft}{SFT}{Supervised Fine-Tuning}
\newacronym{gpu}{GPU}{Graphics Processing Unit}
\newacronym{IEEE}{IEEE}{Institute of Electrical and Electronics Engineers}
\newacronym{WiMAX}{WiMAX}{Worldwide Interoperability for Microwave Access}
\newacronym{ASN}{ASN}{access service network}
\newacronym{ASN-GW}{ASN-GW}{ASN gateway}
\newacronym{CSN}{CSN}{Connectivity Service Network}
\newacronym{oran}{O-RAN}{Open Radio Access Network}
\newacronym{ric}{RIC}{Radio Intelligent Controller}
\newacronym{rt}{RT}{Real-Time}
\newacronym{uav}{UAV}{Unidentified Aerial Vehicle}
\newacronym{v2x}{V2X}{Vehicle-To-Everything}

\newacronym{nwdaf}{NWDAF}{Network Data Analytics Function}
\newacronym{pcf}{PCF}{Policy Control Function}
\newacronym{mtlf}{MTLF}{Model Training Logical Function}
\newacronym{anlf}{ANLF}{Analytics Logical Function}

\newacronym{PA}{PA}{power amplifier}
\newacronym{ecgan}{ECGAN}{Enhanced Capsule Generation Adversarial Network}

\newacronym{NI}{NI}{National Instruments}

\newacronym{TDD}{TDD}{time-division duplex}
\newacronym{FDD}{FDD}{frequency-division duplex}
\newacronym{UDP}{UDP}{User Datagram Protocol}
\newacronym{APP}{APP}{application}
\newacronym{mac}{MAC}{medium access control}
\newacronym{phy}{PHY}{physical}
\newacronym{RLC}{RLC}{radio link control}
\newacronym{sdap}{SDAP}{service data adaptation protocol}
\newacronym{FIFO}{FIFO}{first-in first-out}
\newacronym{CRC}{CRC}{cyclic redundancy check}
\newacronym{SAP}{SAP}{service access point}
\newacronym{FEC}{FEC}{forward error correction}
\newacronym{IF}{IF}{intermediate frequency}
\newacronym{RF}{RF}{radio frequency}
\newacronym{mimo}{MIMO}{multiple-input and multiple-output}
\newacronym{MCS}{MCS}{modulation and coding scheme}

\newacronym{SPC}{SPC}{superposition coding}
\newacronym{SVC}{SVC}{Scalable Video Coding}
\newacronym{GM}{GM}{generic multicasting}
\newacronym{SCM}{SCM}{superposition coded multicasting}
\newacronym{SIC}{SIC}{successive interference cancellation}

\newacronym{st}{ST}{secondary transmitter}
\newacronym{pt}{PT}{primary transmitter}
\newacronym{sr}{SR}{secondary receiver}
\newacronym{pr}{PR}{primary receiver}
\newacronym{su}{SU}{secondary user}
\newacronym{pu}{PU}{primary user}

\newacronym{awgn}{AWGN}{additive white Gaussian noise}

\newacronym{cdf}{CDF}{cumulative density function}
\newacronym{ccdf}{CCDF}{complementary CDF}
\newacronym{iid}{i.i.d.}{independent and identicaly distributed}
\newacronym{rf}{RF}{radio frequency}
\newacronym{hbf}{HBF}{Hybrid Beamforming}

\newacronym{dd}{DD}{Device-to-Device}
\newacronym{ddu}{DDU}{Device-to-Device user}
\newacronym{dds}{DDS}{Device-to-Device system}
\newacronym{ddt}{DT}{DDU transmitter}
\newacronym{ddr}{DR}{DDU receiver}

\newacronym{bs}{BS}{base station}
\newacronym{bsu}{BSU}{base station associated user}
\newacronym{bsas}{BSAS}{base station associated system}
\newacronym{bst}{BT}{BSU transmitter}
\newacronym{bsr}{BR}{BSU receiver}

\newacronym{epg}{EPG}{energy per goodbit}
\newacronym{mepg}{MEPG}{modified energy per goodbit}
\newacronym{ee}{EE}{energy efficiency}
\newacronym{se}{SE}{spectral efficiency}

\newacronym{wrt}{w.r.t.}{with respect to}

\newacronym{kkt}{KKT}{Karush-Kuhn-Tucker}
\newacronym{al}{AL}{Active Learning}
\newacronym{admm}{ADM}{Alternating Directing Method}
\newacronym{cr}{CR}{cognitive radio}
\newacronym{ssi}{SSI}{soft-sensing information}
\newacronym{csi}{CSI}{Channel State Information}
\newacronym{qsi}{QSI}{queue state information}
\newacronym{el}{EL}{enhancement layer(s)}
\newacronym{snr}{SNR}{signal-to-noise ratio}

\newacronym{NAL}{NAL}{network abstraction layer}
\newacronym{QP}{QP}{quantization parameter}

\newacronym{ofdm}{OFDM}{orthogonal frequency-division multiplexing}
\newacronym{ofdma}{OFDMA}{orthogonal frequency-division multiple access}
\newacronym{tdma}{TDMA}{time division multiple access}

\newacronym{PUSC}{PUSC}{partial usage of the subchannels}
\newacronym{CFO}{CFO}{carrier frequency offset}
\newacronym{I/Q}{I/Q}{in-phase and quadrature-phase}
\newacronym{ASK}{ASK}{amplitude-shift keying}
\newacronym{PSK}{PSK}{phase-shift keying}
\newacronym{BPSK}{BPSK}{binary phase-shift keying}
\newacronym{QPSK}{QPSK}{quadrature phase-shift keying}
\newacronym{QAM}{QAM}{quadrature amplitude modulation}
\newacronym{PSNR}{PSNR}{peak signal-to-noise ratio}
\newacronym{PELR}{PELR}{packet error and loss rate}

\newacronym{kNN}{\textit{k}-NN}{\textit{k}-nearest neighbor algorithm}
\newacronym{SVM}{SVM}{support vector machines}
\newacronym{nn}{NN}{neural network}
\newacronym{NN}{NN}{neural network}
\newacronym{dnn}{DNN}{deep neural network}
\newacronym{RBF}{RBF}{radial basis function}
\newacronym{RMSE}{RMSE}{root mean squared error}
\newacronym{mse}{MSE}{mean squared error}
\newacronym{lmse}{LMSE}{linear mean square-error estimator}

\newacronym{R2}{$R^2$}{coefficient of determination}

\newacronym{KAUST}{KAUST}{King Abdullah University of Science and Technology}
\newacronym{GSA}{GSA}{Global mobile Suppliers Association}

\newacronym{VoD}{VoD}{video on demand}
\newacronym{HEVC}{HEVC}{High Efficiency of Video Coding}
\newacronym{DASH}{DASH}{Dynamic Adaptive Streaming over HTTP}

\newacronym{PUT}{PUT}{people using television}

\newacronym{ADTVS}{ADTVS}{Audience Driven live TV Scheduling}

\newacronym{arq}{ARQ}{automatic repeat request}

\newacronym{harq}{HARQ}{hybrid automatic repeat request}

\newacronym{sdp}{SDP}{semi-definite programming}

\newacronym{tcp}{TCP}{transmission control protocol}

\newacronym{e2e}{E2E}{end-to-end}

\newacronym{ran}{RAN}{radio access network}
\newacronym{cran}{CRAN}{cloud radio access network}
\newacronym{udcran}{UD-CRAN}{ultra-dense CRAN}
\newacronym{dran}{DRAN}{distributed radio access network}
\newacronym{hcran}{H-CRAN}{hybrid cloud radio access network}
\newacronym{hetnet}{HetNet}{heterogeneous network}
\newacronym{vcran}{V-CRAN}{virtualized CRAN}
\newacronym{ecran}{E-CRAN}{edge-CRAN}
\newacronym{hvcran}{H-VCRAN}{hybrid-virtualized CRAN}

\newacronym{bbu}{BBU}{baseband processing unit}
\newacronym{rrh}{RRH}{remote radio head}
\newacronym{ru}{RU}{radio unit}
\newacronym{rs}{RS}{remote site}
\newacronym{cs}{CS}{central site}

\newacronym{rru}{RRU}{radio resource unit}
\newacronym{rb}{RB}{resource block}
\newacronym{hpn}{HPN}{high-power node}
\newacronym{lpn}{LPN}{low-power node}
\newacronym{mabs}{MaBS}{macro basestation}

\newacronym{comp}{CoMP}{coordinated multi-point}
\newacronym{ranaas}{RANaaS}{RAN-as-a-Service}

\newacronym{rof}{RoF}{radio over fiber}
\newacronym{wdm}{WDM}{Wavelength Division Multiplexing}
\newacronym{dls}{DLS}{distributed large scale}

\newacronym{qos}{QoS}{quality of service}
\newacronym{qoe}{QoE}{quality of experience}
\newacronym{qee}{QEE}{quality of energy-efficiency}

\newacronym{gg}{GG}{group-to-group}
\newacronym{ht}{HT}{hyper-transceiver}

\newacronym{fh}{FH}{fronthaul}
\newacronym{dl}{DL}{downlink}
\newacronym{ul}{UL}{uplink}

\newacronym{cp}{CP}{Cell-Processing}
\newacronym{up}{UP}{User-Processing}

\newacronym{co}{CO}{center office}

\newacronym{du}{DU}{digital unit}
\newacronym{lc}{LC}{Line-Card}

\newacronym{onu}{ONU}{optical network unit}
\newacronym{olt}{OLT}{optical line terminal}
\newacronym{osw}{OSW}{optical switch}

\newacronym{es}{ES}{ethernet switch}

\newacronym{ppp}{PPP}{Poisson point process}

\newacronym{mppp}{MPPP}{marked Poisson point process}

\newacronym{sinr}{SINR}{signal to noise and interference ratio}

\newacronym{sir}{SIR}{signal to interference ratio}

\newacronym{mbs}{MBS}{macro basestation}
\newacronym{ap}{AP}{access point}
\newacronym{fap}{FAP}{femto-cell access point}
\newacronym{sap}{SAP}{small-cell access point}
\newacronym{iot}{IoT}{Internet of Things}
\newacronym{ti}{TI}{Tactile Internet}
\newacronym{lsm}{LSM}{linear scalarizing method}

\newacronym{lp}{LP}{Low-Priority}
\newacronym{hp}{HP}{High-Priority}
\newacronym{lpu}{LPU}{Low-Priority user}
\newacronym{hpu}{HPU}{High-Priority user}
\newacronym{lps}{LPS}{Low-Priority system}
\newacronym{hps}{HPS}{High-Priority system}

\newacronym{ttm}{TTM}{time to market}
\newacronym{udn}{UDN}{ultra-dense network}

\newacronym{capex}{CAPEX}{capital expenditure}
\newacronym{opex}{OPEX}{operational expenditure}

\newacronym{cpri}{CPRI}{common public radio interface}
\newacronym{otn}{OTN}{optical transport network}
\newacronym{pon}{PON}{passive optical network}
\newacronym{twdm}{TWDM}{time and wavelength division multiplexing}

\newacronym{ec}{EC}{Edge-Cloud}
\newacronym{cc}{CC}{Central-Cloud}

\newacronym{mmw}{m-Wave}{Milli-Meter wave}

\newacronym{gops}{GOPS}{giga operation per second}
\newacronym{mops}{MOPS}{mega operation per second}

\newacronym{ip}{IP}{internet protocol}
\newacronym{rlc}{RLC}{radio link control}
\newacronym{pdcp}{PDCP}{packet data convergence protocol}

\newacronym{mno}{MNO}{mobile network operator}
\newacronym{prb}{PRB}{physical resource block}
\newacronym{mi}{MI}{modulation index}

\newacronym{wifi}{WiFi}{wireless local area network}
\newacronym{cpu}{CPU}{central processing unit}
\newacronym{vcpu}{VCPU}{virtual CPU}
\newacronym{vm}{VM}{virtual machine}

\newacronym{urs}{UrS}{user requested service}

\newacronym{rsf}{RSF}{radio sub-frame}
\newacronym{siso}{SISO}{single-input single-output}
\newacronym{ram}{RAM}{random access memory}
\newacronym{xr}{XR}{Extended Reality}

\newacronym{nef}{NEF}{Network Exposure Function}

\newacronym{vr}{VR}{Virtual Reality}
\newacronym{agv}{AGV}{Automated Guided Vehicle}

\newacronym{rssi}{RSSI}{Received Signal Strength Indicator}

\newacronym{mec}{MEC}{mobile edge computing}
\newacronym{co2}{CO$_{2}$}{carbo dioxide}


\newacronym{cfp}{CFP}{communication function processing}
\newacronym{ptp}{PTP}{precision time protocol}

\newacronym{mdt}{MDT}{Model Drive Test}

\newacronym{voip}{VoIP}{voice over Internet protocol}
\newacronym{sdn}{SDN}{Software Defined Network}

\newacronym{da}{DA}{data analytics}

\newacronym{kpi}{KPI}{key performance indicator}

\newacronym{noc}{NOC}{Network Operations Centre}
\newacronym{fso}{FSO}{Field Service Operations}

\newacronym{fsmc}{FSMC}{finite state markov chain}

\newacronym{5G}{5G}{5th Generation of Mobile Networks}
\newacronym{nr}{NR}{new radio}
\newacronym{gnbcu}{gNB-CU}{gNB central unit}
\newacronym{gnbdu}{gNB-DU}{gNB distributed unit}
\newacronym{ecpri}{eCPRI}{common public radio interface}
\newacronym{fl}{FL}{federated learning}
\newacronym{rsrq}{RSRQ}{Reference Signal Received Quality}
\newacronym{rsrp}{RSRP}{Reference Signal Received Power}
\newacronym{urllc}{URLLC}{ultra-reliable low-latency communications}
\newacronym{embb}{eMBB}{enhanced mobile broadband}

\newacronym{mae}{MAE}{modified autoencoder}
\newacronym{mtc}{MTC}{machine type communication}
\newacronym{mmtc}{mMTC}{massive machine type communication}
\newacronym{pca}{PCA}{principal component analysis}

\newacronym{cps}{CPS}{cyber-physical system}
\newacronym{gnb}{gNB}{gNodeB}
\newacronym{ref}{REF}{reliability enhancement feature}
\newacronym{nfo}{NFO}{network level feature orchestrator}
\newacronym{dc}{DC}{data center}
\newacronym{vnf}{VNF}{virtual network function}
\newacronym{nssmf}{NSSMF}{Network Slice Subnet Management Function}
\newacronym{ai}{AI}{Artificial Intelligence}
\newacronym{rl}{RL}{Reinforcement Learning}
\newacronym{ddpg}{DDPG}{deep deterministic policy gradient}
\newacronym{dqn}{DQN}{deep Q-networks}
\newacronym{sac}{SAC}{soft actor-critic}
\newacronym{a2c}{A2C}{advantage actor-critic}
\newacronym{td3}{TD3}{twin delayed deep deterministic policy gradient algorithm}
\newacronym{poc}{PoC}{proof of concept}
\newacronym{cvae}{CVAE}{Conditional Variational AutoEncoder}
\newacronym{oam}{OAM}{Operation, Administration, and Maintenance}

\newcommand*\myglsentry[1]{%
  \protect\ifglsused{#1}{%
    \glsentryshort{#1}%
  }{%
    \glsentrylong{#1}%
  }%
}

\newacronym{v2n}{V2N}{Vehicle To Network}
\newacronym{v2n2v}{V2N2V}{Vehicle to Network to Vehicle}
\newacronym{fdd}{FDD}{Frequency Division Duplexing}
\newacronym{cots}{COTS}{Commercial Off-The-Shelf}
\newacronym{dme}{DME}{Dedicated Measurement Equipment}
\newacronym{cgan}{CGAN}{Conditional Generative Adversarial Network}
\newacronym{dtpc}{DT-PC}{Digital Twin for Protocol and Connectivity}
\newacronym{jcas}{JCAS}{joint communication and sensing}
\newacronym{aql}{AQL}{Action-conditioned Q-Learning}

\title{Using Large Language Models to Understand Telecom Standards}

\hyphenpenalty=90000
\author{\IEEEauthorblockN{Athanasios Karapantelakis, Mukesh Thakur, Alexandros Nikou, Farnaz Moradi, \\ Christian Olrog, Fitsum Gaim, Henrik Holm, Doumitrou Daniil Nimara, Vincent Huang}
\IEEEauthorblockA{\textit{Ericsson AB, Torshamnsgatan 21, 16483, Stockholm, Sweden} \\
\{athanasios.karapantelakis, mukesh.thakur, alexandros.nikou,  farnaz.moradi, christian.olrog, \\fitsum.gaim.gebre, henrik.holm, doumitrou.nimara, vincent.a.huang\}@ericsson.com}
}

\maketitle

\begin{abstract} 
The Third Generation Partnership Project (3GPP) has successfully introduced standards for global mobility. However, the volume and complexity of these standards has increased over time, thus complicating access to relevant information for vendors and service providers. Use of Generative Artificial Intelligence (AI) and in particular Large Language Models (LLMs), may provide faster access to relevant information. In this paper, we evaluate the capability of state-of-art LLMs to be used as Question Answering (QA) assistants for 3GPP document reference. Our contribution is threefold. First, we provide a benchmark and measuring methods for evaluating performance of LLMs. Second, we do data preprocessing and fine-tuning for one of these LLMs and provide guidelines to increase accuracy of the responses that apply to all LLMs. Third, we provide a model of our own, TeleRoBERTa, that performs on-par with foundation LLMs but with an order of magnitude less number of parameters. Results show that LLMs can be used as a credible reference tool on telecom technical documents, and thus have potential for a number of different applications from troubleshooting and maintenance, to network operations and software product development.
\end{abstract}


\section{Introduction}

The Transformer model architecture, proposed in 2017, enabled faster training of \gls{ml} sequence models than previous architectures, such as \glspl{rnn} and in particular \glspl{lstm} \cite{DBLP:journals/corr/VaswaniSPUJGKP17}. In the field of \gls{nlp}, Transformer-based \glspl{llm} were shown to perform well in text-based applications such as language generation, knowledge utilization, complex reasoning, structured data generation and information retrieval \cite{zhao2023survey}. Applications that use \glspl{llm}, are based in so-called ``foundation models'', such as OpenAI's \gls{gpt} \cite{openai2023gpt4} and Meta's Llama \cite{touvron2023llama2} family of models. Foundation models are designed to generate a wide and general variety of outputs and can be used for standard tasks such as summarization, completion, \gls{qa} and code generation. They have billions of parameters and are trained on corpora such as Common Crawl~\cite{commoncrawl}. 

For foundation models to adapt to application domains, such as telecommunications, prompt engineering and fine-tuning approaches are used. Prompt engineering partially relies on ``in-context learning'', where models learn by using examples as context \cite{dong2023survey}, and partially on guidelines such as patterns and templates for prompt authoring that more accurately describe the task a model should generate outputs for \cite{white2023prompt}. Prompts are provided to models  as ``embeddings'', i.e., numerical representations organized in vectors that indicate commonality between words in prompts and provide a format required for further processing at the \gls{llm} itself.

\begin{figure}[t!]
  \centering
  \includegraphics[width=\linewidth]{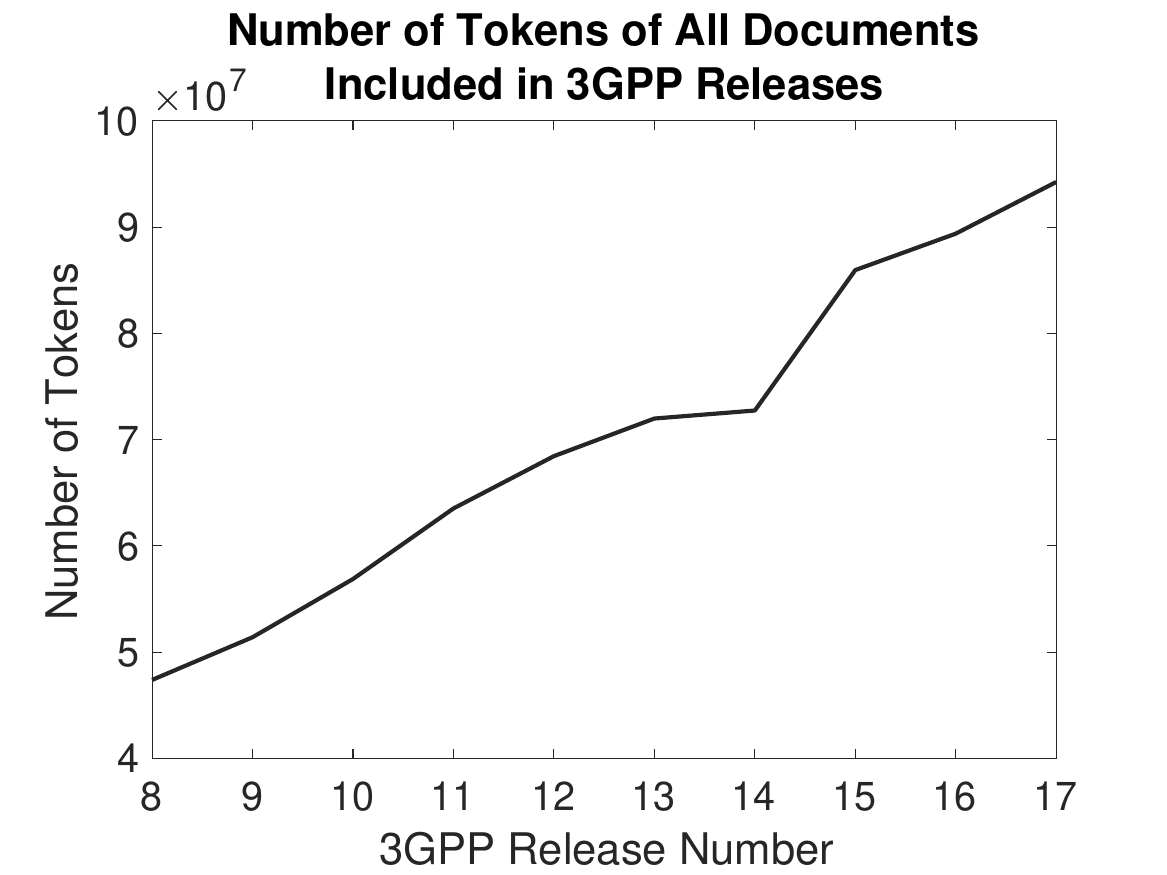}
  \caption{Increase of number of tokens (words) for 3GPP releases, between Release 8 (2006-01-23) and Release 17 (2018-06-15). At the moment of writing of this paper, Releases 18 and 19 still have an ``Open'' status, meaning that material is being added to the documents comprising those Releases.}
  \label{3gpp_complexity}
\end{figure}

While prompt engineering controls the output of a model, fine-tuning can be used to add knowledge in a specific area that may not be covered sufficiently by initial training data. Fine-tuning techniques include offline learning approaches, such as \gls{sft}, which uses an already available labeled dataset, and online learning approaches such as \gls{rlhf}, which uses human preferences as a reward function \cite{wang2023aligning}. 

In this paper, we investigate whether \glspl{llm} can be used as digital assistants for \gls{3gpp} standards. By default, these standards are created to be human-readable and have been increasing in size as newer generations of mobile networks have additional functionality when compared to their predecessors (see Figure~\ref{3gpp_complexity}). This means that it is progressively more difficult for human readers, but also machines, to find relevant information in these standards.

This paper contributes to the state of the art by determining the capabilities and limitations of the current generation of LLM when it comes to referencing 3GPP standards. Given that all telecom standards follow the same natural language document structure, we envision the applicability of this paper to go beyond 3GPP family of standards and also generally apply to other standardization bodies, such as TeleManagement Forum (TMForum), European Telecommunication Standards (ETSI), Open Radio Acess Network (O-RAN) and International Telecommunication Union (ITU).

The contribution of this paper to \gls{soa} is threefold:
\begin{itemize}
\item Evaluation of performance of \gls{soa} foundation \glspl{llm} for \gls{qa} of \gls{3gpp} documents using both statistical and reference metrics.
\item Introduction of TeleRoBERTa, an extractive \gls{qa} model, that performs as well as the highest-performing foundation models, while having fewer parameters.
\item Data preparation methods for \gls{3gpp} documents to reduce hallucinations and increase model perfrormance. Given that telecom standards follow similar structure, methods proposed can be applicable to other families of standards including those published by \gls{tmforum}, \gls{etsi}, \gls{oran} and \gls{itu}.
\end{itemize}

This paper is structured as follows: In section \ref{section2_background}, we describe the background, including related work and determine the gap in \gls{soa} followed by an overview of studies LLMs including TeleRoberta. In section \ref{section3_methodology}, we describe the architecture of the system we built to evaluate the models and describe our methodological approach.  In section \ref{section5_evaluation}, we describe the experiments we carried out to evaluate model quality. The section includes a description of insfrastructure, benchmark dataset and key performance indicators (KPI) used in the assessment, and ends with identifying capabilities and limitations of models. Finally, in section \ref{section6_conclusion}, we conclude by recapitulating our key findings and highlighting areas of future research.

\section{Background} \label{section2_background}

\subsection{Related Work}

\gls{genai} and \glspl{llm} have been successfully utilized specialized domains including telecom. A comprehensive survey on \gls{genai} including \glspl{llm} for mobile networks is presented in~\cite{Karapantelakis2023survey}, identifying several use cases, including customer incident management, configuration management, billing plan generation and exposure to third parties. It is envisioned that \glspl{llm} will open up a new era for wireless networks and LLM-based intelligence will become ubiquitous in the future networks. Several recent studies have looked into the opportunities, application, architectural impacts, and challenges that for generative AI and \glspl{llm} in telecommunication networks~\cite{bariah2023large, huang2023large, chen2023netgpt, huang2023ai}.  

The applicability and effectiveness of existing \glspl{llm} has also been studied in the literature. In~\cite{wang2023making}, the applicability and effectiveness of \glspl{llm} for translating high-level policies and requirements specified in natural language into low-level network \glspl{api} is been explored.      
The authors in~\cite{bariah2023telco} studied the adaptability of \glspl{llm} to telecom domain by fine-tuning a number of \glspl{llm} models (BERT, distilled BERT, RoBERTa, and GPT-2) using 3GPP technical specification data from different working groups. The different models were evaluated on a text classification task to determine the 3GPP specification categories with the corresponding working group. The results indicate the applicability of fine-tuned \glspl{llm} to the telecom domain. While promising, the paper has focused on a single classification task and the generative aspects of the models are not studied for understanding telecom standards.  
Using \glspl{llm} for conversational assistance  and question-answering within the telecom domain has also been explored. In~\cite{soman2023telco}, the capabilities, as well as the limitations of \glspl{llm} for conversational assistance related to enterprise wireless products were analyzed. The authors performed experiments using GPT-4, GPT-3.5, Bard (based on LaMDA) and LlaMA models to address different research questions related to domain \gls{qa}, product \gls{qa}, context continuity, and input perturbations caused by spelling errors. The responses generated by \glspl{llm} were evaluated using subjective metrics including Mean Opinion Score (MOS) and inter-rater agreement. Based on the observations, the authors conclude that publicly available \glspl{llm} without fine-tuning are not usable for enterprise use cases. However, the paper does not evaluate any fine-tuned models.   

In~\cite{holm2021bidirectional}, the authors studied the adaptability of a BERT-like model to the telecom domain. In this study, a small 
model was selected due to its resource and training time efficiency and was pre-trained using general data and then adapted to the telecom domain using \gls{3gpp} technology specification files. The authors also developed a benchmark dataset for telecom-specific \gls{qa} (TeleQuAD) to fine-tune and evaluate the models. 

In~\cite{gunnarsson2021multi}, the authors studied multi-hop neural question answering in the telecom domain by adapting neural open-domain question answering systems. Several benchmark datasets, including TeleQuAD and mTeleQuAD, to evaluate telecom \gls{qa} task were introduced using data which is mainly collected from \gls{3gpp} specifications. 

In~\cite{maatouk2023teleqna}, another benchmark dataset for telecom domain (TeleQnA) was introduced and used for evaluation of GPT-3.5 and GPT-4. The  results suggest the need for specialized foundation models, trained or fine-tuned on thelecom data, to enable answering complex telecom questions. 
        
While the potential of using \gls{nlp} and \glspl{llm} for question answering in telecom domain has been considered before, the evaluation of current generation of \glspl{llm} to reference \gls{3gpp} standards and the impact of fine tuning has not been fully explored before.

\subsection{LLM Architectures}

\glspl{llm} typically refer to Transformer-based language models which have tens, or hundreds, of billions of parameters~\cite{zhao2023survey}. The transformer architecture which was first presented in~\cite{vaswani2017attention} has been widely used in different fields including computer vision and audio processing and has revolutionized the field of \gls{nlp} as the go-to architecture~\cite{lin2022survey}. 
A transformer is a sequence-to-sequence model consisting of encoder and decoder stacks. Each encoder consists of a number of identical layers where each layer has a multi-headed self-attention module and a position-wise fully connected feed forward neural network. The decoder similarly consists of a number of identical layers. The decoder additionally performs multi-head attention over the output of the encoder stack. 

The transformer architecture was adapted by OpenAI and they released the first GPT model (GPT-1) in 2018~\cite{radford2018gpt1}. They showed that unsupervised pre-training of a language model followed by discriminative fine-tuning on specific tasks can lead to large gains. GPT-2~\cite{radford2019gpt2} which has 1.5 billion parameters used a similar architecture with some modifications. GPT-3 was introduced in~\cite{brown2020gpt3} as an autoregressive language model with 175B parameters which was trained with a filtered Common Crawl dataset and was utilized in a few-shot, one-shot, or zero-shot way.  GPT-3 used the same architecture as GPT-2 except that alternating dense and locally banded sparse attention patterns were used in the layers of the transformer. OpenAI has also released a set of GPT-3.5 models which are capable of understanding and generating natural language and code, including GPT-3.5 turbo  which is optimized for chat. The latest LLM released by OpenAI is GPT-4~\cite{openai2023gpt4} which is siad to have about 1.76 trillion parameters and is a multimodal model that can accept image and text inputs and producing text outputs. 

In contrast to GPT LLMs, the Meta's LLaMA models \cite{touvron2023llama1, touvron2023llama2} are open-source, and use data from a mixture of publicly available sources. In July 2023, LLaMA-2~\cite{touvron2023llama2} was released, with versions from 7 to 70 billion parameters. LLaMA-2 has some architectural differences compared to LLaMA-1 including  increased context length and grouped-query attention (GQA). 
 
Falcon is an \gls{llm} with 180 billion parameters which is trained on the RefinedWeb dataset~\cite{penedo2023refinedweb}. The Falcon family of \glspl{llm} is also based on the transformer architecture and uses multiquery attention for improved scalability. At the time of writing, Falcon is the largest openly available \gls{llm} which is shown to outperform LLaMA-2 70B model and GPT-3.5 on Massive Multitask Language Understanding tasks.  

The above models belong to generative \gls{qa} category, wherein new text is generated based on previous knowledge. Another class of models are the extractive \gls{qa} models, wherein text is extracted and quoted as is from prior knowledge. In the extractive question-answering setup,  the encoder type models such as BERT~\cite{bert} and RoBERTa~\cite{roberta} have demonstrated significant progress in the past years. Performing on par with estimated human performance on large-scale benchmarks such as SQuAD \cite{rajpurkar-squad-v1, rajpurkar-squad-v2}. In the context of telecom, TeleRoBERTa \cite{holm2021bidirectional, gunnarsson2021multi} was introduced as adaptation of the RoBERTa base model to the characteristics the telecommunications domain, trained on a large corpus of text collected from in-domain sources such as 3GPP specification. 

Table~\ref{table_llms} summarises the different  \gls{llm} that were considered in this paper.

\begin{table}[t!]
\caption{LLMs used for the experiments } \label{table_llms}
\begin{tabular}{p{2cm}|p{2cm} |c|c} 
\hline
Model &  Creator &License type & Parameters \\
\hline
GPT-3.5-Turbo & OpenAI& Proprietary &  175B  \\
GPT-4 & OpenAI & Proprietary & undisclosed \\
LLaMA-2 & Meta & LLaMA-2.0 & 7B, 13B, 70B \\
Falcon & Technology Innovation Institute & Apache 2.0 & 180B \\
TeleRoBERTa & Ericsson & Proprietary & 124M \\
\hline
\end{tabular}
\end{table}

\section{Methodology} \label{section3_methodology}

This section describes the methods used to design and develop the system along with its architecture. Conventionally, foundation models are pre-trained with large dataset with point-in-time to ensure their effectiveness. However, there may be instances where it becomes necessary to work with one's own/new dataset. In such cases, two approaches can be used: fine-tuning, i.e., further training of the foundation model with the new dataset for increased accuracy of response, within specific domains, and Retrieval Augmented Generation (RAG) \cite{rag2020} with which the user can feed its own dataset to the foundation model and guide the model response in real time through queries (prompts), enhancing the overall model response. In another context, in specific use cases that LLMs have to be loaded to devices with limited resources, quantization methods can be used, as will be described in Section III.B.

\subsection{Retrieval Augmented Generation}

Typically, A RAG system comprises four core components. A document processor that parses different document types and splits a lengthy document into smaller sizes or chunks, which is crucial for enhancing the relevance of content retrieval in response to queries. Additionally, there exists a limit on number of tokens a foundation model can process per operation. Another key component is an embedding model that creates vector representations of the document, which capture the semantic meaning of text, allowing to quickly and efficiently find other pieces of text with similar content. This vector representation created by the embedding model is simply called embedding. Another component is a type of database that can efficiently store and search for these embeddings. This database is called the vectorstore. Lastly, a retriever takes the user's query, which might have context, and semantically searches the vectorstore for similar results. A user query with context is referred to as a prompt. 

Prompts can be zero-shot, where a certain instruction is prepended to the user query without providing the model with any direct examples. This makes sure the model focuses on providing desired answers. Few-shot prompting refers to the case where a few examples are prepened to the user query for a more accurate response. Lastly, chain-of-thought prompting, allows for detailed problem-solving by guiding the models through intermediate steps. In this work, we have used zero-shot prompt to get precise answers.

\begin{figure*}[t!]
 \centering
 \includegraphics[width=\linewidth]{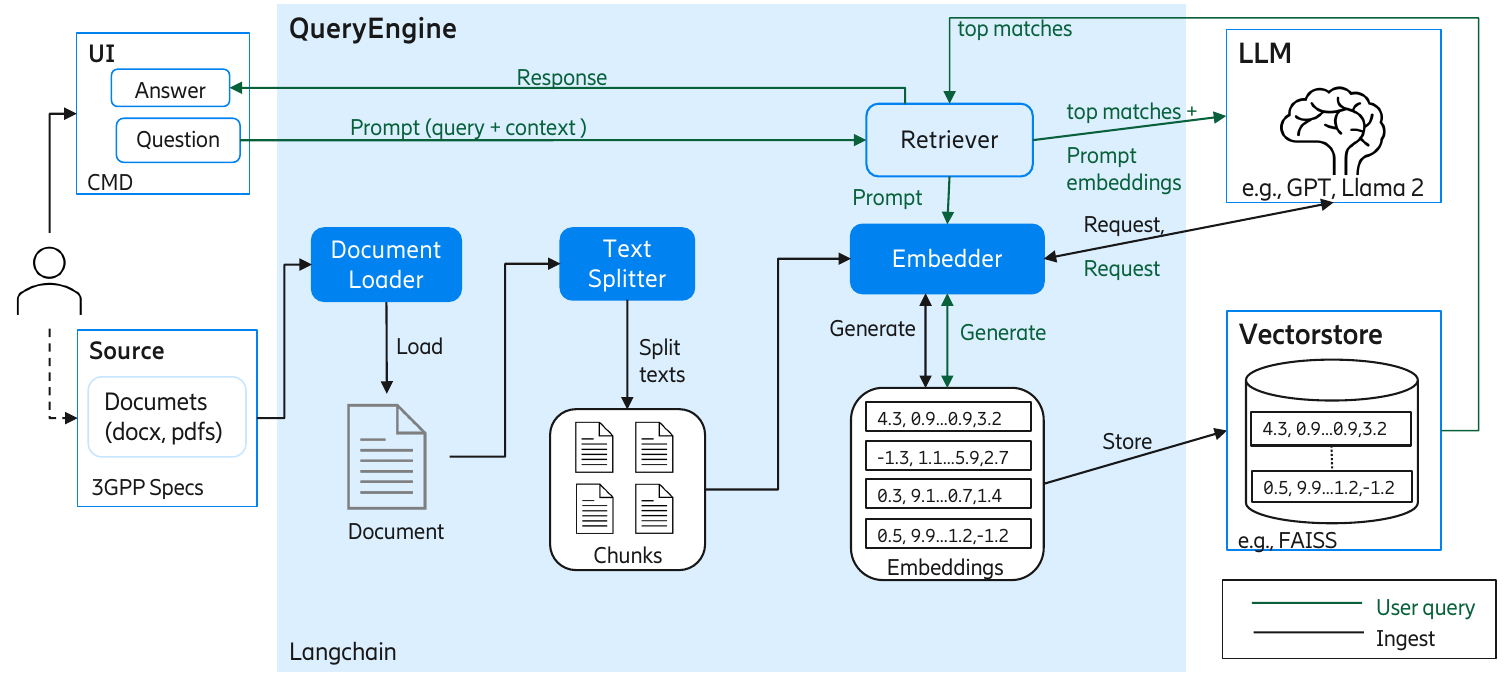}
 \caption{Architecture of TelcoGenAI}
 \label{architecture}
\end{figure*}

\subsection{Quantization methods for LLMs}

In this section, we describe two quantization methods used for faster inference in LLMs. Namely, the Georgi Gerganov’s Machine Learning (GGML) and GPT-Generated Unified Format (GGUF) \cite{ggml}. GGML and GGUF are two quantization methods used in large LLMs to reduce the precision of the model's weights and activations from floating-point numbers to integers. GGML was an early attempt to create a file format for storing GPT models. It allowed models to be shared in a single file, making it convenient for users since it could be run on CPUs, which made them accessible to a wider range of users. On August 21st 2023, GGML was replaced by GGUF which aims to address the limitations of GGML and improve the overall user experience, by offering more flexibility, extensibility, and compatibility with different types of LLMs. For the experimentation of the paper in hand, the GGUF quantization was adopted.

\subsection{Fine-tuning}

For fine-tuning we used the \gls{sft} method, which uses a labelled dataset with domain data in order to guide the model to a specific task. In the context of this work, the task was to provide \gls{qa} reference to \gls{3gpp} standards, therefore the questions revolved around content included in those standards. Of specific interest, is content that can be misinterpreted by the \gls{llm}, such as acronyms or technical jargon, that may also overlap with prior knowledge from another domain that the \gls{llm} was trained with. We discuss more details on the fine-tuning approach in section \ref{subsec:ft}.

\subsection{System Architecture}

In order to interact/query with dataset that is \gls{3gpp} documents, we did fine-tuning of a \gls{llm} model and also designed and developed a system based on the RAG. This system is \textit{TelcoGenAI}. The architecture of \textit{TelcoGenAI} is illustrated in Figure~\ref{architecture}. The main component of the \textit{TelcoGenAI} is \textit{QueryEngine}. This component is based on Langchain~\cite{langchain2023} framework, an open source framework, designed to create LLM applications. The \textit{QueryEngine} utilizes four main langchain features: document loader, text splitter, embedder and retriever \cite{langrag2023}. The document loader, loads the 3GPP specification word/pdf documents. The text splitter, splits the loaded documents into designated chunks sizes (e.g., 1000) with overlap (e.g., 100). Overlap ensures there is context between chunks. The embedder ingests these chunks and creates embeddings using specified LLM, subsequently storing them to one of the popular vectorstore, Facebook AI Similarity Search (FAISS) \cite{faiss2023}. The retriever, accepts a user query along with context forming a prompt and generates prompt embeddings using the specified LLM. These embeddings are used to search the vectorstore for the most relevant matches. The top matches and prompt embeddings both, is then sent to the LLM, which generates a response for the user.  

\section{Evaluation} \label{section5_evaluation}

\subsection{Introduction} \label{section5_evaluation_subsec:intro}

In this section, we present our experiments and results. The models we used for our experiments are illustrated in table \ref{table_llms}. With the exception of GPT 3.5-Turbo, which uses OpenAI's \gls{api}, all other models were downloaded and experimented with locally. The reason for selecting to experiment with the aforementioned models out of a plurality of models was twofold. First, we wanted to have a mixture of models of varying complexity, that could potentially be executed in either standard off-the shelf servers, or require dedicated \gls{gpu} clusters, to address different use-cases and budget. Second, we wanted to have models that could be executed locally as well as models that could be interfaced with remotely. The former could potentially allow prompting or fine-tuning with proprietary information that cannot be disclosed to a third-party infrastructure, via an \gls{api}. 

We have completed two sets of experiments:

\begin{itemize}
    \item The first set is an evaluation of selected \gls{soa} foundation models of varying complexity, as well as the TeleRoBERTa model, using a benchmark we developed in-house known as TeleQuAD \cite{holm2021bidirectional}. The benchmark is one of the first datasets for \gls{3gpp} standards and contains over 4,000 question and answer pairs that can be used for evaluating performance of \glspl{llm}. The foundation models were prompted with a template, and embedded \gls{3gpp} specifications as context.
    \item The second set uses learnings from the first set to construct a labeled dataset from both \gls{3gpp} specification and other sources, such as content from web sites. This dataset was then used to further train a foundation model using the \gls{sft} approach.
\end{itemize} 

To measure performance, we use two types of metrics. First, BERTScore \cite{bert-score}, which measures similarity of generated answers of \glspl{llm} with TeleQuAD. The process includes transforming the natural language of generated and reference answers to embeddings in vector space using a pretrained embedding model. Next, these vectors are compared using a pairwise cosine-similarity approach. Second, we use GPT-4 to evaluate generated answers of \glspl{llm}, using the \textit{QAEvalChain} functionality of langchain. GPT-4 is arguably the most capable \gls{llm} to date, and was used to verify whether correctness of the \gls{llm}-generated answer  by comparing it against a reference answer from TeleQuAD. We refer to this metric as ``GPT-4 Ref score''.

\subsection{Evaluation of Foundation Model Performance} \label{section5_subsection_evaluation_foundation_LLM}

\begin{figure}[t!]
  \centering
  \includegraphics[width=\linewidth]{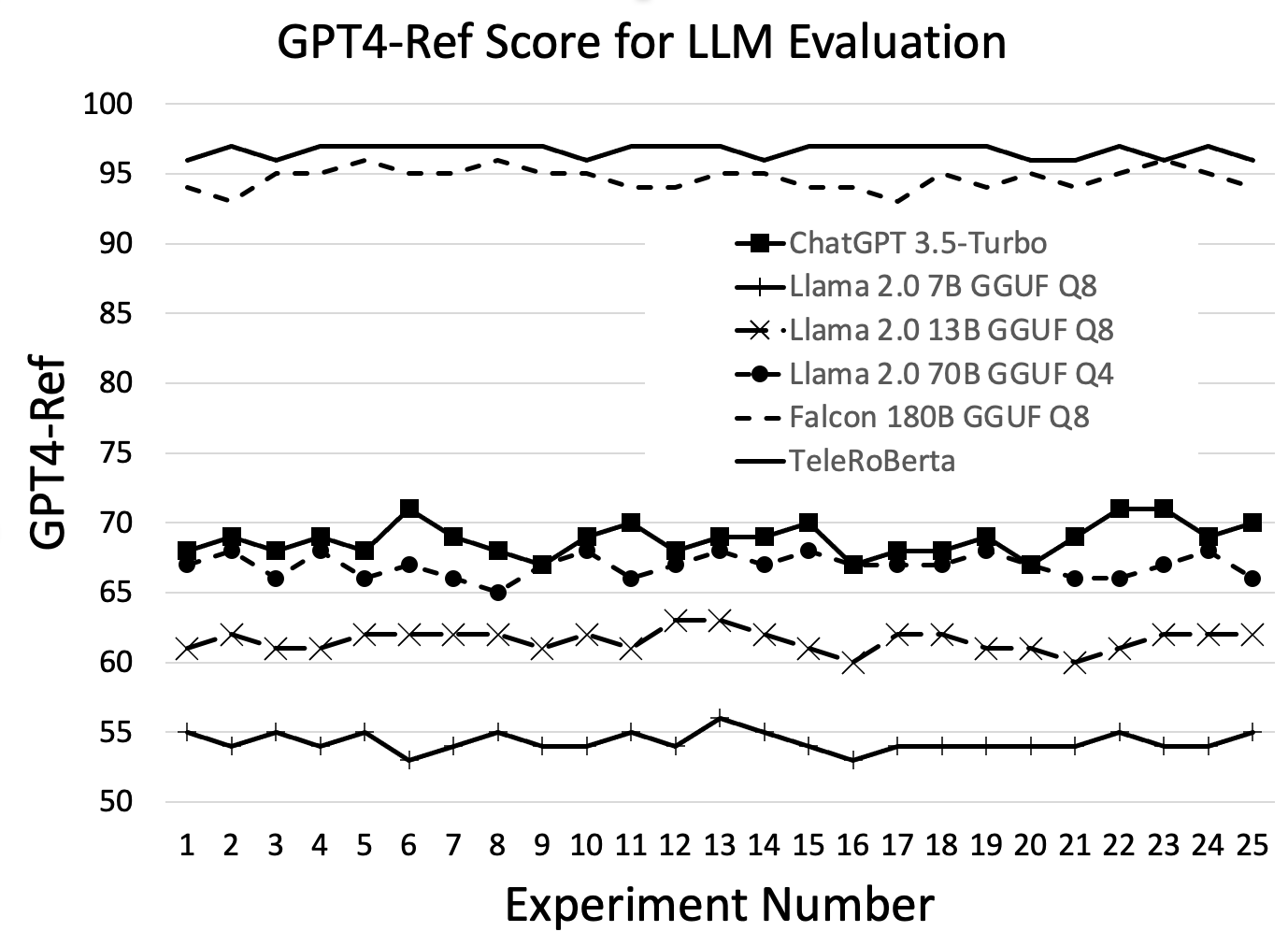}
  \caption{Performance of foundation \glspl{llm}, using TeleQuAD dataset as ground truth and GPT-4 as reference.}
  \label{fig:gpt4ref_scores}
\end{figure}

For this set of experiments, we evaluated the performance of the five \glspl{llm} mentioned in section \ref{section5_evaluation_subsec:intro} against TeleQuAD. For local models, we used C++ ports instead of the original model, and 8-bit (Q8) and 4-bit (Q4) introduced under the GGUF format by llama.cpp \cite{llamacpp}. The advantage of using such an approach was that we were able to perform inference on Llama and Falcon model variants on commodity hardware, using system \gls{ram} instead of \gls{gpu} memory and without sacrificing model performance. In order to verify performance consistency, the experiments were performed for every \gls{llm} 24 times. Figure \ref{fig:gpt4ref_scores} shows the results for GPT-4 Ref.

\begin{figure}[t!]
  \centering
  \includegraphics[width=\linewidth]{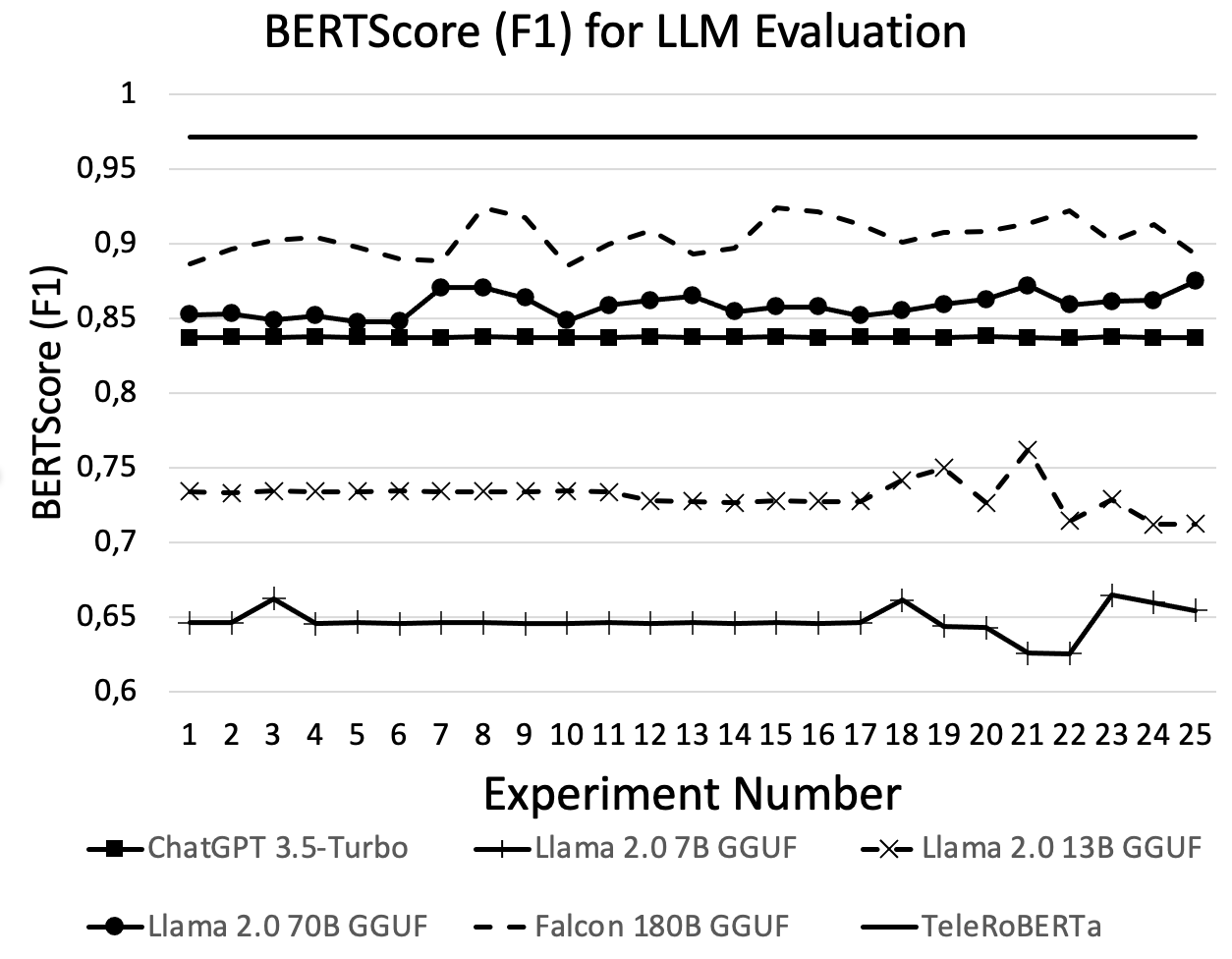}
  \caption{Performance of foundation \glspl{llm}, using BERTScore and TeleQuAD dataset as ground truth.}
  \label{fig:bertscore}
\end{figure}

We observe that TeleRoBERTa and Falcon 180B performed similarly, answering 92 to 96\% of all questions correctly, whereas models with lower number of parameters had lower performance. These measurements were also mirrored when using BERTScore, as shown in Figure \ref{fig:bertscore}. The small differences can be attributed to the way the BERTSCore is calculated, which is based on the distance between tokens of answers, which can themselves be based on semantics between words (such as categories they belong to, to establish relative meaning). On the other hand, GPT-4 relies on the large corpus of knowledge it was trained with to make that decision. Another observation is that the statistical dispersion among 24 runs is small, as illustrated in Figures \ref{fig:bertscore} and \ref{fig:gpt4ref_scores}. This means that the performance figures for the \glspl{llm} are consistent and therefore can be trusted.

For comparison, we evaluated the TeleRoBERTa model using the machine reading comprehension approach \cite{rajpurkar-squad-v1}. Given a context paragraph and a question, the model predicts the answer span in the context. This is a different formulation with how the generative \glspl{llm} were evaluated, and yields insightful results for comparison.
Before evaluation, the model is fine-tuned on a combination of SQuAD-v2 and TeleQuAD with training samples of 100K and 3K, respectively. The training follows a similar approach as discussed in \cite{bert, roberta}.
We observe that the model performs on par or better than the much larger models, such as Llama and GPT-3.5 Turbo, while having only 125M trainable parameters. This is an indication that domain-adapted smaller models can be very competitive on specific tasks.
However, it is important to recognize that the benchmark used in this work was annotated for the extractive QA, and hence verbatim answer spans from context will inherently match closer to the ground truth answers.
Furthermore, the generative \glspl{llm} tend to phrase the answers with variance and extra explanations that may affect the computed scores. For benchmarks that require multi-source content aggregation and complex reasoning, the generative \glspl{llm} are expected to perform better than their encoder counter parts -- this exploration is left for future work.

When examining the incorrect responses from different \glspl{llm}, we found some common denominators that pointed to potential issues with regards to how the embedded context was represented, but also lack of potential background knowledge. Specifically we noticed the following:

\begin{itemize}
    \item Technical jargon and abbreviations were  misinterpreted by \glspl{llm}, leading them to provide answers that were irrelevant to the meaning of the question. This phenomenon is also known as ``bias''. An example was a question about \gls{hss}, which was interpreted by an \gls{llm} as ``Home Security Solution''. 
    \item Information organized in tables was particularly hard for \gls{llm} to identify. For example, several technical standards have a list of what are the mandatory and optional parameters for \gls{api} requests. When querying the model whether a parameter was mandatory or optional, the network could not identify the cell in the table were this parameter was located, and provide a seemingly random answer that differed accross inferencing sessions. Such phenomena, where an \gls{llm} seemingly ``guesses'' the information due to lack of knowledge or its inability to process knowledge, is known as ``hallucinations''.
    \item Cross-reference of information in documents was also difficult for \glspl{llm} to follow. For example, such information may be contained as reference to another \gls{3gpp} standard or another source of information,
\end{itemize}

In the next section, we discuss the approach we took to work around these issues. 

\subsection{Context Engineering and Fine-tuning} \label{subsec:ft}

\begin{figure}[t!]
  \centering
  \includegraphics[width=\linewidth]{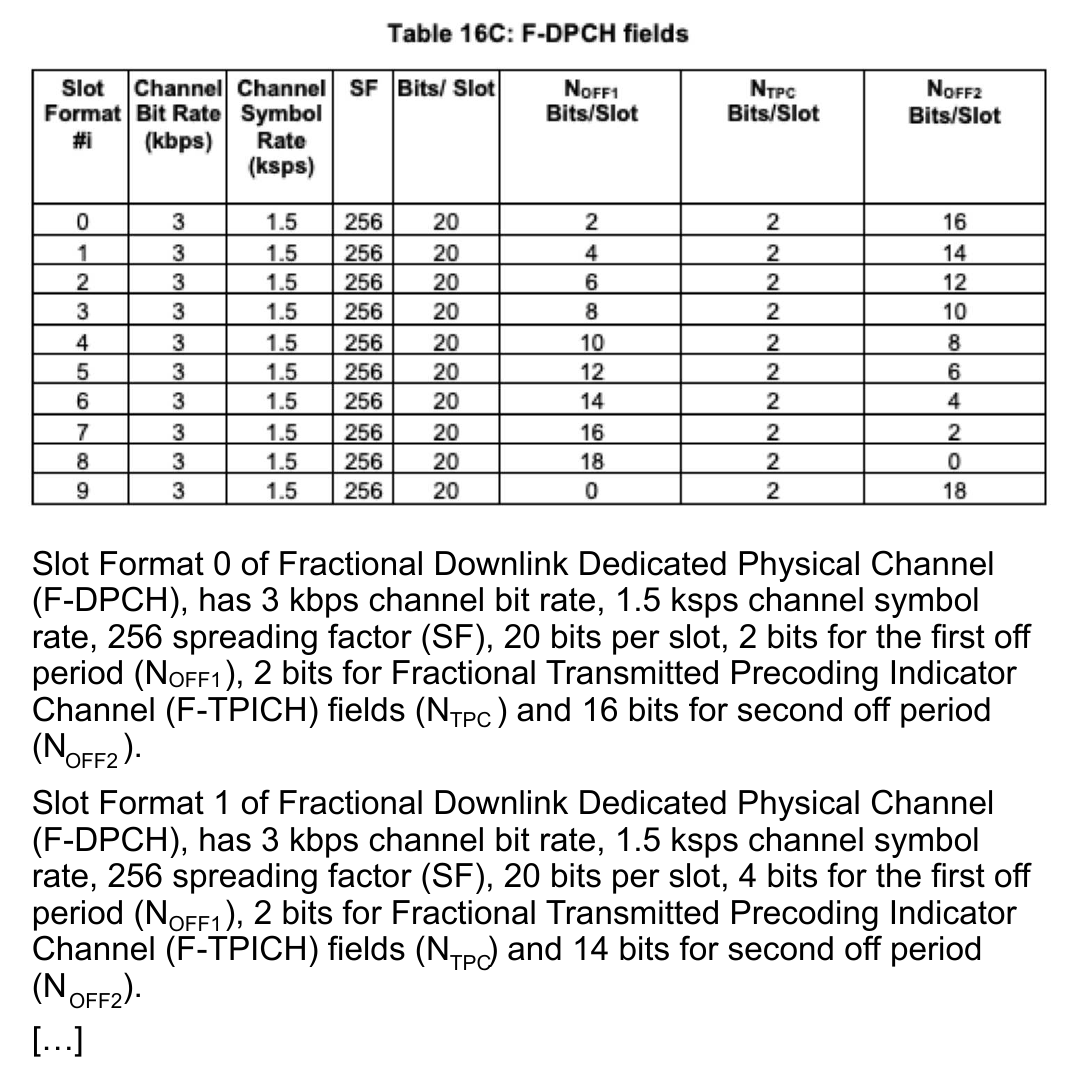}
  \caption{Table transformation: Exemplary table from a \gls{3gpp} specification (top) and replacement text (bottom).}
  \label{fig:preprocessing}
\end{figure}

In order to address the issues mentioned in section \ref{section5_subsection_evaluation_foundation_LLM}, we carried a small-scale experiment on a small subset of TeleQuAD dataset. This subset contained a number of \gls{qa} pairs that demonstrated these issues. Prior to fine-tuning, we prepared the \gls{3gpp} documents used as context. Specifically, we replaced the tables with natural language. We have also replaced the abbreviations in text with an explanation of the abbreviation. This was done to reduce the bias mentioned in the previous section. Figure \ref{fig:preprocessing} illustrates an example. 

In addition, we included all documents referenced by \gls{3gpp} documents as context to help reduce hallucinations. To reduce bias, we also performed fine-tuning. Specifically, we created a labeled dataset of question and answer pairs comprising information that could be overlapping with irrelevant data that the foundation model was originally trained with. For example, we asked simple questions such as ``What is HSS?'', and answer with ``Home Subscriber Server'', in order to help guide the model towards the telecom definition of HSS. For fine-tuning we used LlaMA 2 with 7B parameters and trained it using \gls{sft}. Results are shown in table \ref{table_finetuning}. The BERTScore and GPT-4 Ref score are averages of 24 runs of the three models on the limited test dataset, hence the slight diversion from the data presented in Figures \ref{fig:gpt4ref_scores} and \ref{fig:bertscore}.

\begin{table}[t!]
\caption{Performance of fine-tuned model versus baseline} \label{table_finetuning}
\begin{tabular}{p{1cm}p{3.5cm}lc} 
Model Name & Method & BERTScore   & GPT-4 Ref \\
\hline
Llama 2.0 7B                   & Prompted, Contexted with raw data                  & 0,674382848 & 56        \\
Llama 2.0 13B                  & Prompted, Contexted with raw data                  & 0,774348293 & 65        \\
Llama 2.0 7B                   & Fine-tuned, prompted, contexted with processed data & 0,783238483 & 64     \\  
\hline
\end{tabular}
\end{table}

The results from the evaluation show an approximate 16\% increase in performance of the fine-tuned and properly contexted model, over the baseline LlaMA~2 7B. These performance figures are on-par with the baseline of LlaMA~2 13B, which also means that \glspl{llm} with smaller number of parameters - and thus lower computational and storage requirements can offer comparable levels of performance when fine-tuned to larger \glspl{llm}.

\section{Conclusion} \label{section6_conclusion}

In this paper we investigate the capabilities and limitations of state-of-the-art \glspl{llm} as \gls{qa} assistants for telecom domain. 

To assist users to access relevant information faster from ever growing information source of 3GPP specifications, we introduce \textit{TelcoGenAI}, a platform that provies access to difference LLMs.

We also introduce \textit{TeleRoBERTa}, an extractive \gls{qa} \gls{llm}, and compare its performance with the state-of-art foundation generative \gls{qa} \glspl{llm}, such as GPT 3.5-Turbo. For comparison, we use \textit{TeleQuAD}, a benchmark containing \gls{qa} pairs based on \gls{3gpp} standard content. We use two types of metrics to measure accuracy of produced answers, namely BERTScore, and GPT-4 Ref. Whereas the former method is based on statistical evaluation of distance between the reference and produced answer to a question, the latter is based on the use of GPT-4, arguably the highest performing LLM, to evaluate whether a produced answer is similar to the reference answer. 

Results not only show that \textit{TeleRoBERTa} performs on-par with the state-of-art foundation \glspl{llm} that have an order of magnitude more parameters, but also that the accuracy is consistently high enough for these \glspl{llm} to be used as credible digital assistants for \gls{3gpp} standards reference. Results also show that through pre-processing of prompt context and use of \gls{sft}, accuracy can be further improved.

Establishing a baseline set of \glspl{llm} that perform well in \glspl{3gpp} specification \gls{qa} opens the way for many interesting applications, from field service operations such as troubleshooting, commissioning and upgrading of radio base stations, to customer incident management at a \gls{noc} and is part of future work. In addition, as the number of APIs in mobile networks keeps increasing, given that functionality is further compartmentalized and virtualized, software development-related tasks such as genreation of API calls is an interesting area to explore.

\section*{Acknowledgment}
The authors would like to acknowledge Dr. Ioanna Mitsioni for her review of the paper prior to submission.

\bibliographystyle{IEEEtran}
\bibliography{ref}
\end{document}